\documentclass[5p]{elsarticle}

\usepackage{lineno,hyperref}
\modulolinenumbers[5]

\usepackage{times}
\usepackage{epsfig}
\usepackage{graphicx}
\usepackage{amsmath}
\usepackage{amssymb}  
\usepackage{stfloats}
\usepackage{bm}
\usepackage{multirow}
\usepackage{latexsym}
\usepackage{threeparttable}
\usepackage{xcolor}
\usepackage{footmisc}
\usepackage{makecell}
\usepackage{bbding}
\usepackage{framed}
\usepackage{subfigure}
\usepackage{booktabs}
\usepackage{algorithm}
\usepackage{algorithmic}
\usepackage{tablefootnote}
\usepackage{tabularx} 
\usepackage{pifont} 
\usepackage{microtype} 
\usepackage{nicefrac} 
\usepackage{url}
\usepackage{colortbl}

\journal{Pattern Recognition}

\let\oldequation\equation
\let\oldendequation\endequation

\renewenvironment{equation}
{\linenomathNonumbers\oldequation}
{\oldendequation\endlinenomath}

\begin{document}

\hypersetup{
    colorlinks=false,
    pdfborder={0 0 0}
}

\begin{frontmatter}

\title{MegaHan97K: A Large-Scale Dataset for Mega-Category Chinese Character Recognition with over 97K Categories}

\author[mymainaddress,secaddress]{Yuyi Zhang$^{\dag}$}
\ead{yuyi.zhang11@foxmail.com}

\author[mymainaddress]{Yongxin Shi$^{\dag}$}
\ead{yongxin_shi@foxmail.com}

\author[mymainaddress]{Peirong Zhang}
\ead{eeprzhang@mail.scut.edu.cn}

\author[mymainaddress]{Yixin Zhao}
\ead{yixin_zhao01@126.com}

\author[mymainaddress]{Zhenhua Yang}
\ead{eezhyang@gmail.com}

\author[mymainaddress,secaddress]{Lianwen Jin$^*$}
\ead{lianwen.jin@gmail.com}


\address[mymainaddress]{School of Electronic and Information Engineering, South China University of Technology, Guangzhou, China.}
\address[secaddress]{SCUT-Zhuhai Institute of Modern Industrial Innovation, Zhuhai, China}

\begin{abstract}

Foundational to the Chinese language and culture, Chinese characters encompass extraordinarily extensive and ever-expanding categories, with the latest Chinese GB18030-2022 standard containing 87,887 categories. 
The accurate recognition of this vast number of characters, termed mega-category recognition, presents a formidable yet crucial challenge for cultural heritage preservation and digital applications.
Despite significant advances in Optical Character Recognition (OCR), mega-category recognition remains unexplored due to the absence of comprehensive datasets, with the largest existing dataset containing merely 16,151 categories. 
To bridge this critical gap, we introduce MegaHan97K, a mega-category, large-scale dataset covering an unprecedented  97,455 categories of Chinese characters. 
Our work offers three major contributions: (1) MegaHan97K is the first dataset to fully support the latest GB18030-2022 standard, providing at least six times more categories than existing datasets; (2) It effectively addresses the long-tail distribution problem by providing balanced samples across all categories through its three distinct subsets: handwritten, historical and synthetic subsets; (3) Comprehensive benchmarking experiments reveal new challenges in mega-category scenarios, including increased storage demands, morphologically similar character recognition, and zero-shot learning difficulties, while also unlocking substantial opportunities for future research.
To the best of our knowledge, the MetaHan97K is likely the dataset with the largest classes not only in the field of OCR but may also in the broader domain of pattern recognition.
The dataset is available at \href{https://github.com/SCUT-DLVCLab/MegaHan97K}{https://github.com/SCUT-DLVCLab/MegaHan97K}.

\end{abstract}





\end{frontmatter}

\renewcommand{\thefootnote}{\fnsymbol{footnote}}
\footnotetext[2]{Co-first author.}
\footnotetext[1]{Corresponding authors.}
\renewcommand{\thefootnote}{\arabic{footnote}}
\section{Introduction}
\label{intro}


Optical Character Recognition (OCR) is a fundamental task in computer vision that has garnered intensive research attention for decades~\cite{zhou2014minimum, luo2019moran, fang2021read, Open, review}. Within this field, Chinese character recognition (CCR)~\cite{SU2003635, zhang2017online, Z_Wang_Writer, HUANG2022Hipp} has long been a pivotal issue due to its unique challenges.
Unlike Latin, the Chinese character lexicon is continuously expanding to encompass a broader spectrum of infrequently-used and archaic characters. 
For instance, the latest Chinese GB18030-2022~\cite{GB18030_2022} standard includes 87,887 categories, marking a substantial expansion from the 27,533 categories in the preceding GB18030-2000~\cite{GB18030_2000}. 
This expansion reflects increasing demands in areas such as historical document research, digitization of ancient texts, and various social applications\footnote{\href{https://en.wikipedia.org/wiki/GB_18030}{https://en.wikipedia.org/wiki/GB\_18030}}. 
Accurately recognizing mega-category characters poses a significant challenge within the field of CCR and holds paramount significance for the aforementioned areas, impacting not only academic research but also the broader ability to preserve and interact with cultural heritage digitally.
However, to the best of our knowledge, the mega-category challenge remains unexplored, primarily due to data deficiency across several aspects. 
First, although numerous datasets have been released~\cite{hwdb, ahcdb, hccdoc} to propel the advancement of CCR, the majority of them lean to encompass common characters, with only a minor proportion for rare characters. 
Second, the largest dataset to date, M$^5$HisDoc~\cite{m5}, containing only 16,151 categories, is significantly fewer than the total categories in the contemporary Chinese lexicon. 
Third, these datasets suffer from pronounced long-tail distribution issues, with many categories having only one or two samples. 
The paucity of data hampers both the training and evaluation of existing CCR systems, critically stalling the explorations of mega-category CCR. 

\begin{figure*}[t]
  \centering
  \includegraphics[width=0.8\textwidth]{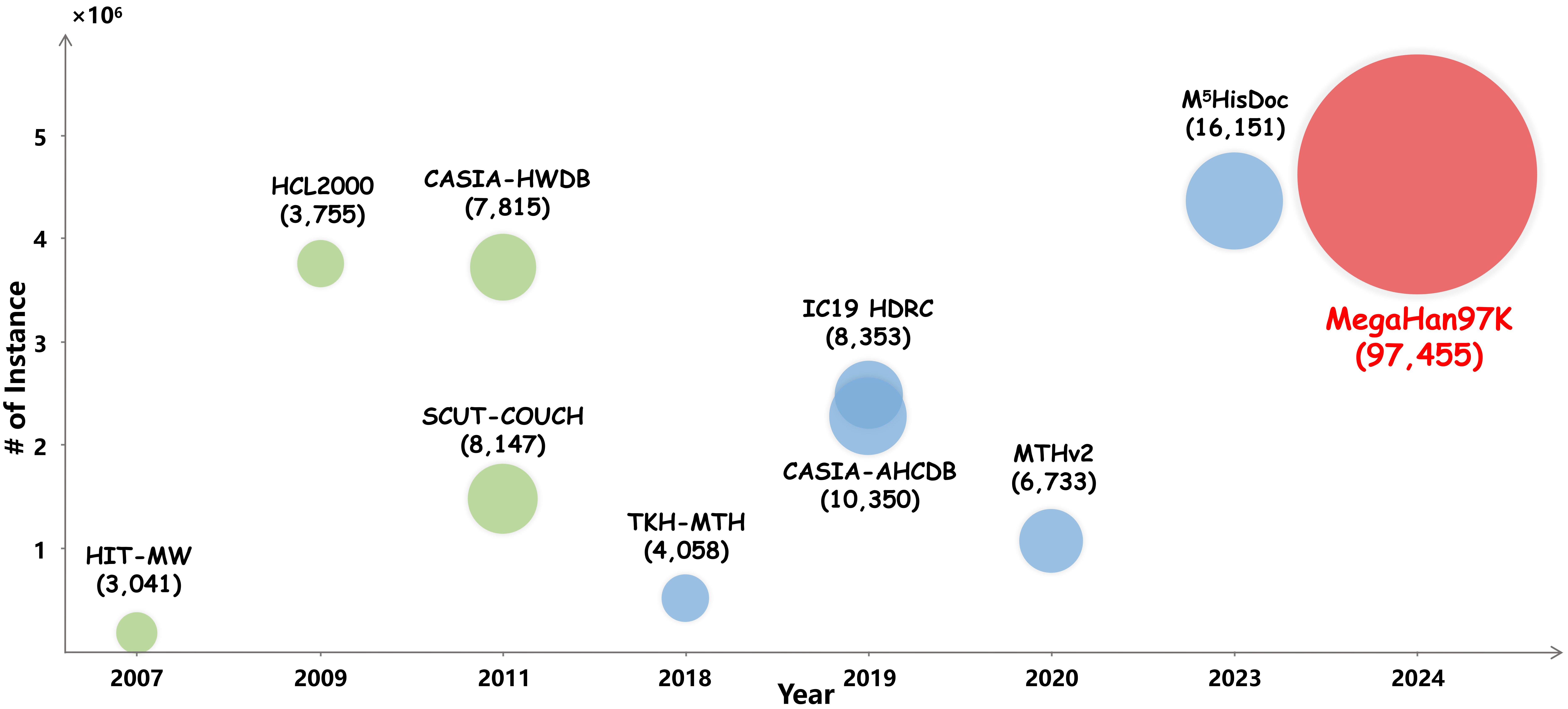}
  \caption{Comparing MegaHan97K with existing Chinese character datasets. The green and blue bubbles represent handwritten and historical datasets, respectively. The area of the bubble and the number in brackets denote categories in each dataset.}
  \label{fig::overview}
\end{figure*}

To fill this gap and advance the development of CCR toward cultural heritage preservation, digital applications, and societal needs, we introduce MegaHan97K, a mega-category, large-scale dataset that contains an unprecedented 97,455 character categories. Figure~\ref{fig::overview} presents a comparison of the MegaHan97K dataset with previous datasets across three dimensions: category, instance count, and data type. 
MegaHan97K offers several key features as follows:
\begin{itemize}
    \item MegaHan97K includes Chinese characters of 97,455 categories, which significantly surpasses existing datasets by at least six times in categories.
    \item MegaHan97K is the first to support the latest Chinese GB18030-2022 standard, ensuring the most comprehensive coverage and compatibility with modern Chinese processing systems. 
    \item MegaHan97K contains three distinct subsets: handwritten, historical, and synthetic. Each subset contains a greater number of character categories compared to existing datasets, resulting in remarkable scale and diversity advantages. 
    \item MegaHan97K effectively mitigates long-tail distribution issues by providing a balanced and sufficient number of samples for each category, ensuring robust training and validation of CCR models.
\end{itemize}

To evaluate the effectiveness of MegaHan97K, we thoroughly benchmark our dataset using various CCR methods, including the state-of-the-art CCR-CLIP~\cite{ccr_clip}, HierCode~\cite{hiercode}, PCSS~\cite{PCSS}, SideNet~\cite{sidenet}, etc. 
The experimental results of the mega-category on MegaHan97K reveal significant challenges for CCR models in this context, such as numerous morphologically similar characters, severe radical zero-shot issues, and increased storage demands.
Furthermore, we conduct cross-validation with previous datasets to show the necessity of a mega-category Chinese characters dataset to drive further progress in this field.

\section{Related Work}
\label{related}

\subsection{Chinese Character Datasets}
\label{relate_dataset}
Existing Chinese character datasets can be mainly divided into two types: handwritten and historical datasets. 
Handwritten datasets tend to focus solely on common Chinese character categories. 
For instance, HCL2000~\cite{hcl2000} comprises 3,755 frequently-used Chinese characters written by 1,000 writers. 
HIT-MW~\cite{hit} includes 853 forms and 3,041 categories authored by over 780 writers. 
SCUT-COUCH2009~\cite{SCUT-COUCH2009} consists of 11 subsets with larger vocabularies and more writers. 
CASIA-HWDB~\cite{hwdb} contains approximately 3.9 million isolated character samples across 7,356 categories, written by 1,020 writers. 
Compared to handwritten datasets, historical counterparts typically encompass a broader range of categories.
TKH-MTH~\cite{mthv1} includes 1,500 images of Chinese historical documents across two types and 4,058 character categories. 
MTHv2~\cite{mthv2}, an extension of TKH-MTH, comprises 3,199 images across the same two types and 6,733 character categories. 
ICDAR 2019 HDRC-CHINESE~\cite{ic19} is a large structured dataset containing 8,353 character categories. 
CASIA-AHCDB~\cite{ahcdb} contains over 2.2 million character samples of 10,350 categories, which are tailored for character recognition in historical documents.
M$^5$HisDoc~\cite{m5} is currently the largest dataset of historical documents, comprising two subsets, five stylistic properties, and 16,151 character categories.
Additionally, CTW~\cite{ctw} is a challenging dataset collected from street views, containing 812,872 Chinese character images spanning 3,650 classes, with complex backgrounds and diverse fonts.
However, these datasets confront several shortcomings. First, no existing dataset supports mega-category scenarios, thereby hindering the progress in related research areas. Second, these datasets still exhibit long-tail distribution problems, with many categories having only one or two samples, which renders them insufficient for effectively training and evaluating CCR models. 
Consequently, there is a pressing need to introduce a new dataset to address these limitations. 

\subsection{Chinese Character Recognition Method}
Early Chinese Character Recognition (CCR) methods~\cite{mcdnn_cnnbase, hccr_cnnbase} employ Convolutional Neural Networks (CNNs) to extract features from character images and achieve exceptional performance. 
However, the challenge of data shortage is exacerbated by the extensive and ever-growing lexicon of Chinese characters, coupled with the labor-intensive data collection and annotation. 
Hence, numerous zero-shot CCR methods~\cite{dmn, wang_2017, prototype_liu_pr_2022, ccd, prototype_liu_pr_2023, icassp_liu_2024} have been proposed to address this issue. 
For instance, DenseRAN~\cite{denseran}, FewShotRAN~\cite{fewran}, and CUE~\cite{rie} consider the recognition task as a radical sequence prediction problem. 
HDE~\cite{hde} and HierCode~\cite{hiercode} encode the Ideographic Description Sequences (IDS) of Chinese characters to construct unique representations for each character. 
Chen et al.~\cite{sld} decomposed Chinese characters into strokes and recognized Chinese characters by predicting stroke sequences. 
SideNet~\cite{sidenet} and ACPM~\cite{mmcharacter} simultaneously leverage the various intrinsic information of Chinese characters, e.g., radicals, strokes, and glyphs, to recognize Chinese characters, achieving favorable results. 
Yu et al.~\cite{ccr_clip} simulated humans recognizing, through aligning the character image and IDS, demonstrating robust zero-shot capabilities. 
Ao et al.~\cite{PCSS} employed a GAN-based model to generate unseen data, thereby significantly enhancing the training and testing processes.
Although existing research methods have demonstrated their zero-shot capabilities on the CASIA-HWDB~\cite{hwdb} and CTW~\cite{ctw} datasets, these validations are typically limited to the scope of 3,755 Level-1 commonly-used characters. 
To date, due to the lack of data, few studies have attempted to evaluate the zero-shot performance of these methods across a broader range of categories.

\section{MegaHan97K Dataset}
\label{megahan}

\subsection{Lexicon Determination}
\label{sec: category}

\begin{figure}[t]
  \centering
  \includegraphics[width=0.95\linewidth]{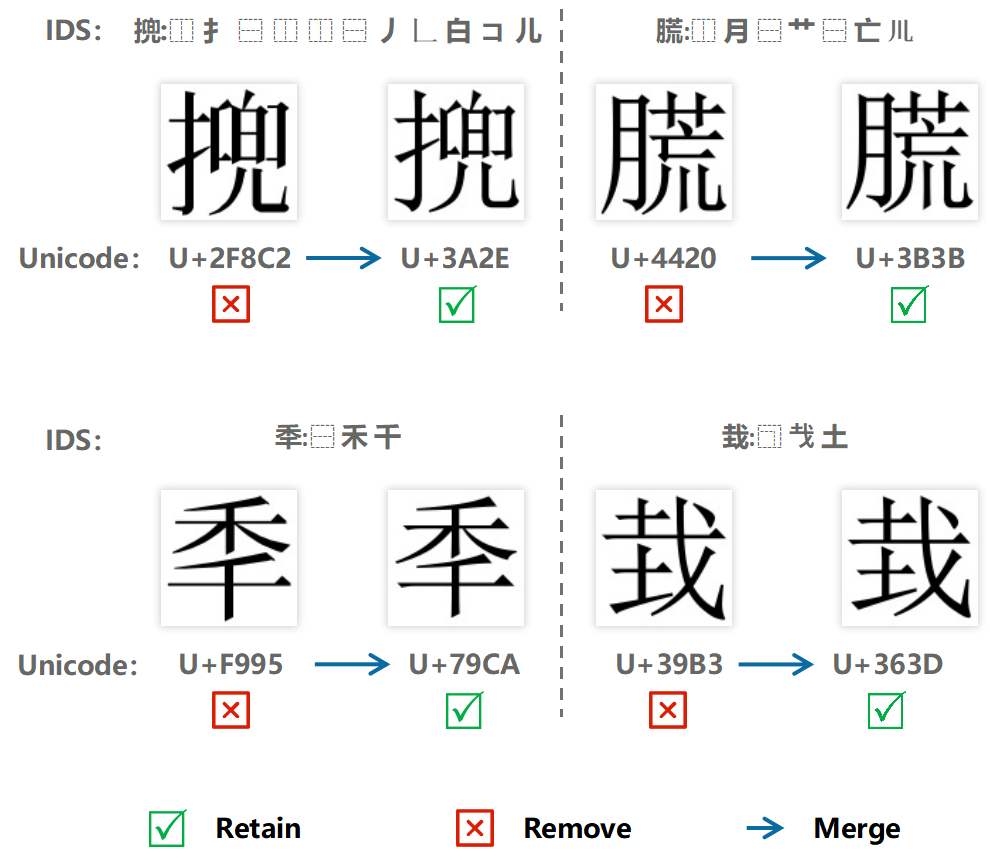}
  \caption{The example of ambiguous Chinese characters.
  }
  \label{fig:ambiguous_characters}
\end{figure}

To meet the escalating demands, the Chinese character lexicon is continuously expanding to encompass a broader spectrum of infrequently-used and archaic characters.
For instance, the Chinese GB18030 standards have shown substantial growth over the years: GB18030-2000~\cite{GB18030_2000}, GB18030-2005~\cite{GB18030_2005}, and GB18030-2022~\cite{GB18030_2022} standards comprise 27,533, 70,244, and 87,887 categories, respectively. 
The Unicode standard\footnote{\href{https://home.unicode.org/about-unicode/}{https://home.unicode.org}} encompasses about 100,000 categories of Chinese characters, exhibiting a broader repertoire that fully covers the GB18030-2022 standard~\cite{GB18030_2022}.
To ensure the broadest possible coverage of the Chinese character, we first incorporate all Chinese characters from GB18030-2022 into our lexicon, then supplement it with Unicode 15.0.0, amassing a total of 98,208 categories of Chinese characters. 
However, we identify certain ambiguous pairs of Chinese characters, which share identical structures and radical formations despite bearing distinct Unicode encodings, as illustrated in Figure~\ref{fig:ambiguous_characters}. 
To avoid confusion about the recognition of such characters, we propose a checking and removing strategy based on IDS comparison. 
Specifically, by comparing the IDS of all characters, we find pairs of characters with the same IDS and represent them uniformly with the smaller Unicode encodings. 
With this strategy, we end up retaining 97,455 Chinese character categories. 

\subsection{Data Acquisition}
\label{sec::acquisition}
MegaHan97K consists of three subsets: historical, handwritten, and synthetic. 
In the following, we will delve into the details of the data acquisition.

\begin{figure}[h]
  \centering
  \includegraphics[width=0.85\linewidth]{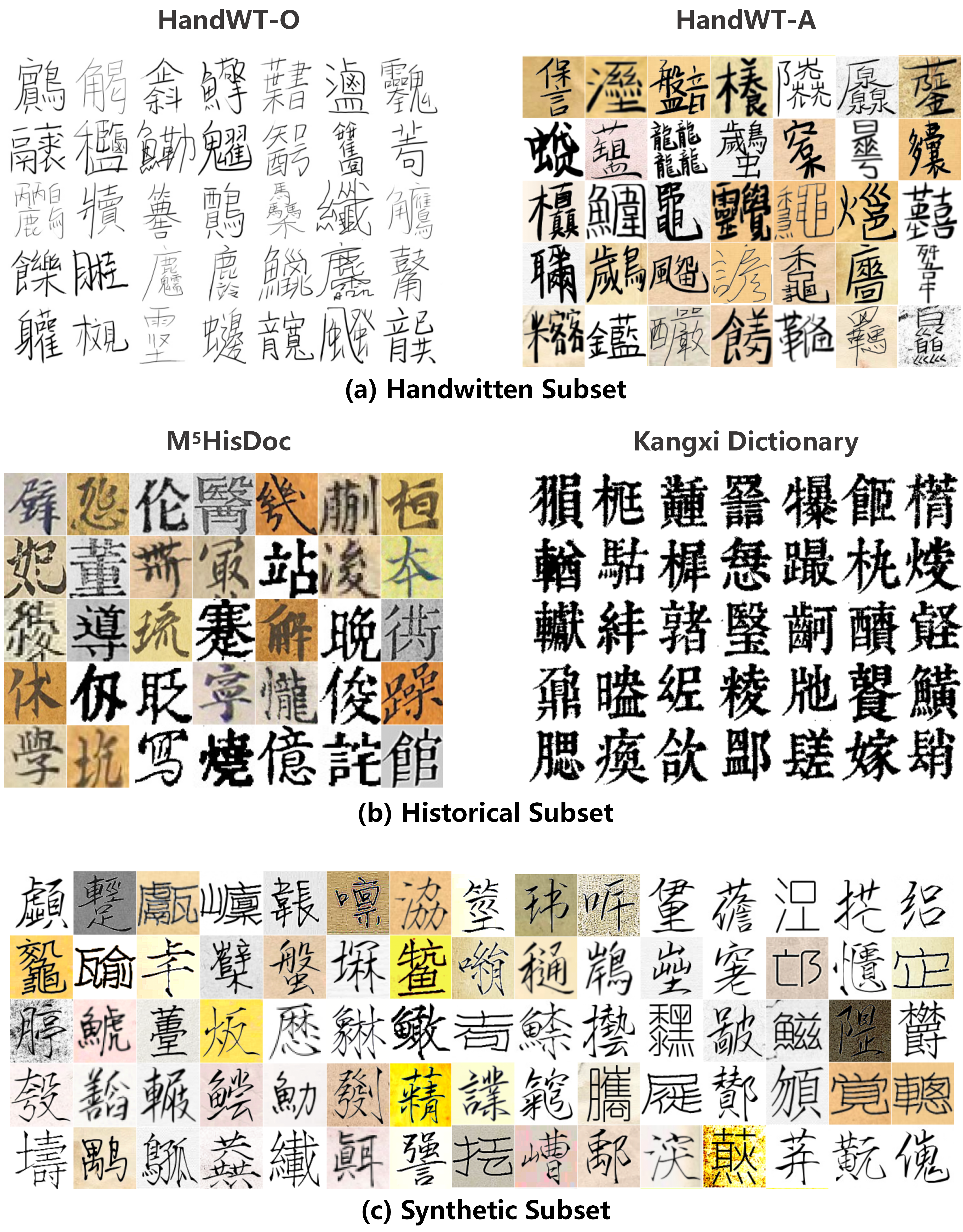}
  \caption{The visualization of the MegaHan97K dataset. HandWT-O (Original) and HandWT-A (Augmented) represent the pre-processed and post-processed versions of the handwritten subset, respectively. Zoom in for better view.}
  \label{fig:all_data}
\end{figure}

\textbf{Historical subset. }
Given the extensive number of rare and variant forms in Chinese characters, samples of these categories are extremely scarce. Considering that most available samples may have already been incorporated into existing datasets, we collect data from M$^5$HisDoc~\cite{m5}, which currently contains the largest variety of character categories, and complement it with additional samples from the HanDian website\footnote{\href{https://www.zdic.net}{https://www.zdic.net}}.
For M$^5$HisDoc, we gather characters based on their sample numbers. 
Specifically, we randomly select 30 samples from the category with more than 30 samples; otherwise, all available samples are included. 
However, M$^5$HisDoc contains only 16,151 Chinese character categories and demonstrates a severe long-tail distribution problem (20\% of the categories have less than 3 samples), making it inadequate for our requirements.
To address these limitations, we turn to the Internet for additional data.
The HanDian website contains carefully organized Chinese character images from historical documents.
Therefore, we develop a web crawler program to download the images and corresponding characters of the \textbf{Kangxi Dictionary} as images and labels to construct the Kangxi Dictionary part in the historical subset. 
The Kangxi Dictionary part encompasses 47,064 Chinese character categories with at least one image per category. 
In total, we have collected approximately 400K samples in the historical subset.
The visualization of the historical subset is presented in Figure~\ref{fig:all_data}(a).

Although we have made extensive efforts to collect historical samples across diverse character categories, the current data exhibits three significant constraints: the category count remains substantially below the 97,455 classes defined in the latest Chinese GB18030-2022 standard, limited samples per category, and pronounced long-tail distribution. To overcome these deficiencies, we propose expanding the dataset through the collection of handwritten Chinese characters.

\begin{figure*}[t]
  \centering
  \includegraphics[width=0.6\linewidth]{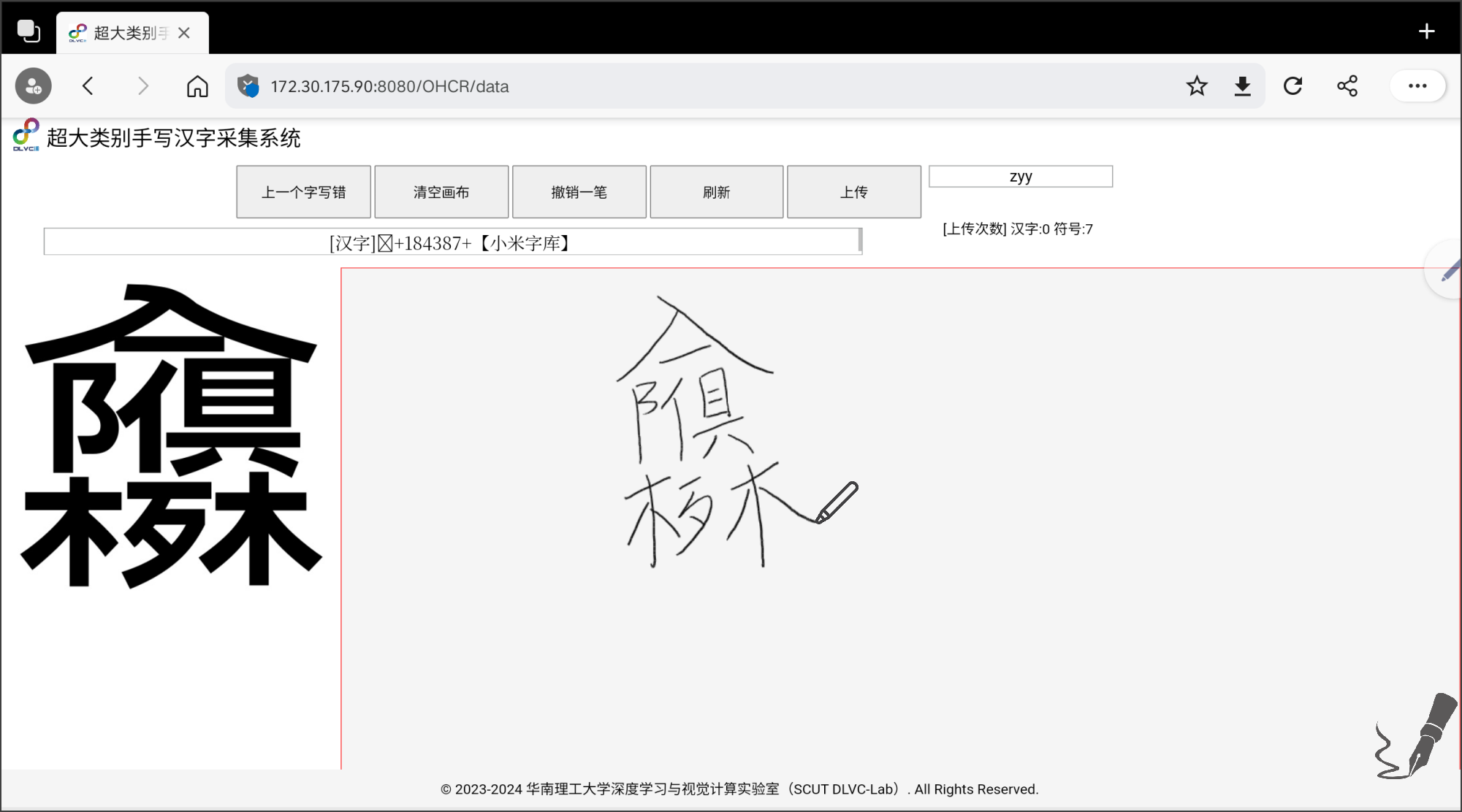}
  \caption{The user interface of the data acquisition website.} 
  \label{fig::interfaces}
\end{figure*}

\textbf{Handwritten subset.}
The handwritten subset constitutes the most critical parts in MegeHan97K, which covers all 97,455 categories of our lexicon. 
For data acquisition, we develop a dedicated website featuring a user interface that includes a main writing board, toolbar, printed character display area, and information display block, as depicted in Figure~\ref{fig::interfaces}.
To enrich data diversity regarding writing mode and writing style, we invite 94 volunteers to write Chinese characters on our website.  
Each volunteer is tasked with writing on Android-based tablets using their specific stylus. 
For the 27,533 categories defined by the GB18030-2000 standard, we collect 20 samples per category. 
Additionally, for the 69,922 categories not included in the GB18030-2000 standard, we collect 5 samples per category. 
In total, we have collected approximately 900K samples.
To ensure the data quality, each sample is verified by at least one of the paper's authors. 
The complete data writing and checking process takes approximately 2,300 person-hours. 

\begin{figure*}[t]
  \centering
  \includegraphics[width=0.7\textwidth]{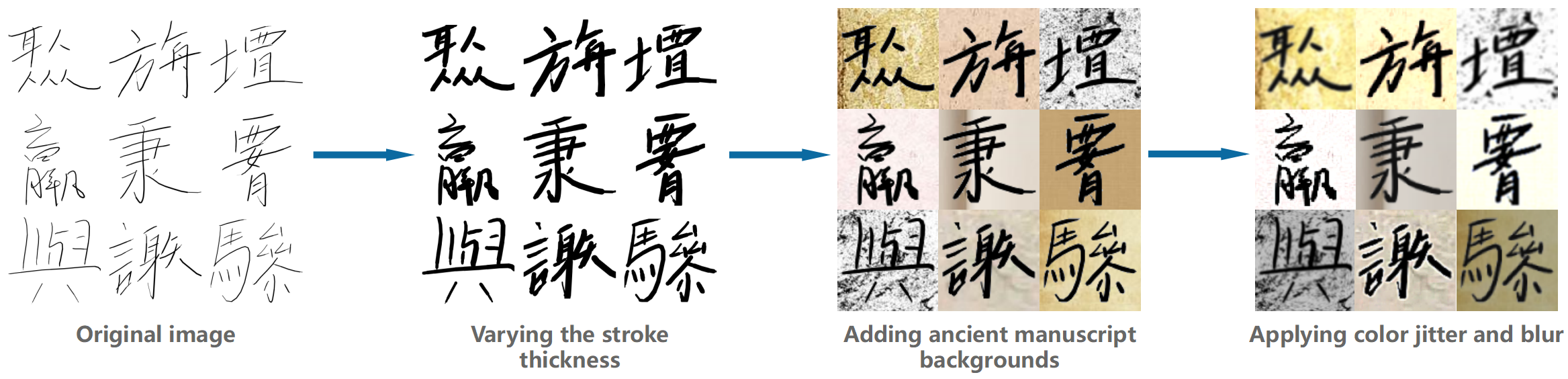}
  \caption{Illustration of data processing to simulate realistic scenarios.}
  \vspace{-5mm}
  \label{fig:data_process}
\end{figure*}

After that, partial clean images undergo processing to simulate characteristics of real-world historical documents as depicted in Figure~\ref{fig:data_process}.
First, we randomly increase the stroke thickness to emulate the style of characters in ancient manuscripts. Second, we choose 30 typical types of ancient manuscript backgrounds from M$^5$HisDoc~\cite{m5} and utilize them to replace the white background of the image randomly. 
Finally, we apply random blurring and color jitter to 80\% of the images while keeping the remaining portion unchanged. 
We denote the pre-processed and post-processed data as \textbf{HandWT-O} (Original) and \textbf{HandWT-A} (Augmented), respectively. 
Notably, the original and augmented parts are sourced from different real data, respectively.
Visualizations of the handwritten subset are presented in Figure~\ref{fig:all_data}(b).

\textbf{Synthetic subset. }
While the aforementioned subsets provide a plentiful total number of samples, the average sample count per category remains constrained, merely sufficient for model validation. 
Drawing inspiration from prior works~\cite{sld, ccr_clip}, we seek to enrich our training data by rendering character images. 
Nevertheless, these works utilizing TrueType Font (TTF) files to render character images are limited to handling a few thousand categories, and merely a handful of TTF files (fewer than five) support the extensive range of 100,000 categories. 
Consequently, we employ the advanced font generation model FontDiffuser~\cite{FontDiffuser} to synthesize character data. 
Specifically, we meticulously select 319 TTF files\footnote{\href{https://www.maoken.com/all-fonts-imgs}{https://www.maoken.com/all-fonts-imgs}} with styles similar to ancient manuscripts and handwritten characters to serve as style image inputs for FontDiffuser, with the PuHuiTi TTF file\footnote{\href{https://www.alibabafonts.com/\#/font}{https://www.alibabafonts.com/\#/font}} employed as the reference content inputs. All employed TTF files are under an open license.
FontDiffuser is trained to mimic the style of these style images while preserving the content of the reference images.
After training, the model is then employed to generate 35 samples in different styles for each category of Chinese characters. 
We extracted 10\% of the synthesized data for sampling inspection to verify that it meets predefined quality standards. 
The explanation for generating 35 samples per category is detailed in Section~\ref{ablation}. 
To simulate real-world scenarios, we apply the augmentation process to all synthetic data, as we implement it on the handwritten subset. 
The visualization of the synthetic subset is shown in Figure~\ref{fig:all_data} (c).

\begin{table}[t]
\caption{A summary of the proposed MegaHan97K dataset. 
`Original' and 'Augmented' represent the original and augmented versions of the handwritten subset, respectively. 
`Kangxi' denotes the Kangxi Dictionary. 
}
\label{tab::summary}
\centering
\renewcommand{\arraystretch}{1.0} 
\resizebox{1.0\linewidth}{!}{
\begin{tabular}{lrrrrrrrrrr}
\specialrule{1.0pt}{1pt}{1pt}
\multirow{2}{*}{Subset}         & \multicolumn{2}{c}{Handwritten} &  & \multicolumn{2}{c}{Historical} & \multirow{2}{*}{Synthetic} & \multirow{2}{*}{Total}   \\ \cline{2-3} \cline{5-6} 
           & Original  & Augmented &  & M$^5$HisDoc & Kangxi &    &     \\  \hline
\#Category       & 97,455         & 96,362    &           & 16,151   & 47,064            & 97,455     & 97,455     \\ 
\#Instance        & 481,455        & 416,486      &        & 353,296  & 49,438            & 3,314,000 & 4,614,675 \\ 
\specialrule{1.0pt}{1pt}{1pt}
\end{tabular}
}
\end{table}

\begin{table}[t]
\small
    \caption{Comparing MegaHan97K with existing Chinese character datasets. `$^\dagger$' indicates that we count only those subsets related to the Chinese GB2312-80 standard in SCUT-COUCH2009. `*' denotes that our analysis is limited to the training set because only the training set of ICDAR 2019 HDRC-CHINESE is available. `Hw.', `His.', and `Syn.' represent handwritten, historical, and synthetic types of data, respectively.}
    \label{tab::dataset_compare}
    \centering
    \renewcommand{\arraystretch}{1.0} 
    \resizebox{1.0\linewidth}{!}{
    \begin{tabular}{lcccccrr}
    \specialrule{1.0pt}{1pt}{1pt}
        \multirow{2}{*}{Dataset} &  \multirow{2}{*}{Year} & \multicolumn{3}{c}{Type}   &  \multirow{2}{*}{\#Instance}    & \multirow{2}{*}{\#Category} \\ \cline{3-5}
           &  &  Hw.  & His.  & Syn. & &    \\ \hline
        HIT-MW~\cite{hit}      & 2007        & \ding{51}  &     &            & 186,444   & 3,041 \\
        HCL2000~\cite{hcl2000} & 2009         & \ding{51}  &     &          & 3,755,000 & 3,755 \\
        SCUT-COUCH~\cite{SCUT-COUCH2009}$^\dagger$ & 2011    & \ding{51}  &   &         & 1,472,420 & 8,147 \\
        CASIA-HWDB~\cite{hwdb}    & 2011     & \ding{51} &   &     & 3,721,874  & 7,815 \\ 
        TKH-MTH~\cite{mthv1}      & 2018     &   & \ding{51}    &              & 521,377  & 4,058 \\
        IC19 HDRC~\cite{ic19}$^*$ & 2019     &   & \ding{51}   &            & 2,482,994  & 8,353 \\
        CASIA-AHCDB~\cite{ahcdb}  & 2019     &   & \ding{51}   &          & 2,276,740  & 10,350 \\
        MTHv2~\cite{mthv2}    & 2020         &   & \ding{51}   &           & 1,081,663   & 6,733  \\
        M$^5$HisDoc~\cite{m5}  & 2023        &   & \ding{51}   &           & 4,367,360   & 16,151 \\ \hline
        MegaHan97K (Ours)     & 2024   & \ding{51}  & \ding{51} & \ding{51}     & 4,614,675  & 97,455 \\ 
        \specialrule{1.0pt}{1pt}{1pt}
    \end{tabular}
    }
\end{table}

\begin{table}[t]
\caption{Category distribution and sample counts of MegaHan97K in general and zero-shot settings. 
 `*' denotes that all data presented are real. }
\label{tab:dataset_div}
\centering
\renewcommand{\arraystretch}{1.0} 
\resizebox{0.95\linewidth}{!}{
\begin{tabular}{lrrrrr}
\specialrule{1.0pt}{1pt}{1pt}
\multirow{2}{*}{Dataset}     & \multicolumn{2}{c}{General Setting}   &    & \multicolumn{2}{c}{Zero-Shot Setting*} \\ \cline{2-3} \cline{5-6}
      & \#Category &  \#Instance  &  & \#Category          & \#Instance           \\ \hline
Training Set & 94,755   &  3,927,457 & & 27,533            & 771,389            \\ 
Test Set*     & 97,455   &  687,218 &  & 69,922            & 404,097            \\  \hline
Total          & 97,455   &  4,614,675 & & 97,455            & 1,175,486         \\ 
\specialrule{1.0pt}{1pt}{1pt}
\end{tabular}}
\end{table}

\begin{table}[t]
\caption{Category distribution and sample counts in the MegaHan97K training set under the general setting. 
`Original' and `Augmented' represent the original and augmented versions of the handwritten subset, respectively. 
`Kangxi' denotes the Kangxi Dictionary. 
}
\label{tab::general_CCR_train}
\centering
\renewcommand{\arraystretch}{1.0} 
\resizebox{1.0\linewidth}{!}{
\begin{tabular}{lcccccccccc}
\specialrule{1.0pt}{1pt}{1pt}
\multirow{2}{*}{Subset} & \multirow{2}{*}{Synthetic}  & \multicolumn{2}{c}{Handwritten} & & \multicolumn{2}{c}{Historical} & \multirow{2}{*}{Total}     \\  \cline{3-4} \cline{6-7} 
     &    & Original & Augmented & & M$^5$HisDoc & Kangxi &         \\ \hline
\#Category & 97,455   & 27,533          & 27,533         & & 9,369 & -   & 97,455    \\
\#Instance  & 3,314,000  & 192,675         & 192,675        & & 228,107 & - & 3,927,457 \\
\specialrule{1.0pt}{1pt}{1pt}
\end{tabular}
}
\end{table}

\subsection{Dataset Analysis}
\label{sec::analysis}
The summarization of the MegaHan97K dataset and the comparison with other datasets are demonstrated in Table~\ref{tab::summary} and~\ref{tab::dataset_compare}, respectively. 
The comparison between MegaHan97K and existing datasets reveals the following advantages. 
First, MegaHan97K encompasses an unprecedented 97,455 categories of Chinese characters, which surpasses existing datasets by at least six times in categories. 
This propels the potential and advancements in the mega-category CCR.
Second, MegaHan97K is the first dataset to completely cover the latest Chinese GB18030-2022 standard, ensuring comprehensive coverage and compatibility with modern Chinese character processing systems. 
Third, MegaHan97K contains three subsets: handwritten, historical, and synthetic. Each contains more character categories than existing datasets, manifesting remarkable scale and diversity advantages. 
Finally, while several datasets feature over ten thousand categories, they often suffer from long-tail distribution issues. 
For instance, approximately 20\% of the categories in M$^5$HisDoc~\cite{m5} have fewer than three samples. 
In contrast, MegaHan97K effectively mitigates long-tail distribution issues by providing a balanced and sufficient number of samples for each category, ensuring robust training and validation of CCR models. The mitigation of the long-tail distribution problem will be further examined in Section~\ref{sec::long-tail}.

\begin{table}[t]
\caption{Category distribution and sample counts in the MegaHan97K test set under the general setting.
`Original' and `Augmented' represent the original and augmented versions of the handwritten subset, respectively. 
`Kangxi' denotes the Kangxi Dictionary. 
}
\label{tab::general_CCR_test}
\centering
\renewcommand{\arraystretch}{1.0} 
\resizebox{1.0\linewidth}{!}{
\begin{tabular}{lcccccccccc}
\specialrule{1.0pt}{1pt}{1pt}
\multirow{2}{*}{Subset} & \multirow{2}{*}{Synthetic}  & \multicolumn{2}{c}{Handwritten} &  &  \multicolumn{2}{c}{Historical} & \multirow{2}{*}{Total}     \\  \cline{3-4} \cline{6-7} 
  &   & Original & Augmented     & & M$^5$HisDoc & Kangxi &         \\ \hline
\#Category & - & 96,362    & 97,455      & & 16,151 & 47,064   & 97,455    \\
\#Instance  & - & 288,780 & 223,811       & & 125,189 & 49,438 & 687,218 \\
\specialrule{1.0pt}{1pt}{1pt}
\end{tabular}
}
\end{table}

\begin{table}[t]
\caption{Category distribution and sample counts in the MegaHan97K training set under the zero-shot setting.
`Original' and `Augmented' represent the original and augmented versions of the handwritten subset, respectively. 
`Kangxi' denotes the Kangxi Dictionary. 
}
\label{tab::zs_train}
\centering
\renewcommand{\arraystretch}{1.0} 
\resizebox{1.0\linewidth}{!}{
\begin{tabular}{lccccccccc}
\specialrule{1.0pt}{1pt}{1pt}
\multirow{2}{*}{Subset}   & \multicolumn{2}{c}{Handwritten} & &  \multicolumn{2}{c}{Historical} & \multirow{2}{*}{Total}     \\  \cline{2-3} \cline{5-6} 
      & Original & Augmented & & M$^5$HisDoc & Kangxi &         \\ \hline
\#Category     & 27,533    & 27,533   & & 8,005 & 22,588   & 27,533    \\
\#Instance      & 275,250   & 275,250  & & 197,562 & 23,327 & 771,389 \\
\specialrule{1.0pt}{1pt}{1pt}
\end{tabular}
}
\end{table}

\begin{table}[t]
\caption{Category distribution and sample counts in the MegaHan97K test set under the zero-shot setting.
`Original' and `Augmented' represent the original and augmented versions of the handwritten subset, respectively. 
`Kangxi' denotes the Kangxi Dictionary. 
}
\label{tab::zs_test}
\centering
\renewcommand{\arraystretch}{1.0} 
\resizebox{0.9\linewidth}{!}{
\begin{tabular}{lccccccccc}
\specialrule{1.0pt}{1pt}{1pt}
\multirow{2}{*}{Subset}   & \multicolumn{2}{c}{Handwritten} &  &  \multicolumn{2}{c}{Historical} & \multirow{2}{*}{Total}     \\  \cline{2-3} \cline{5-6} 
  &    Original & Augmented     &     & M$^5$HisDoc & Kangxi &         \\ \hline
\#Category  & 68,829    & 69,922     & & 1,364 & 24,476   & 69,922    \\
\#Instance  & 206,205 & 141,236     & & 30,545 & 26,111 & 404,097 \\
\specialrule{1.0pt}{1pt}{1pt}
\end{tabular}
}
\end{table}

\section{Experiments}
\label{sec::exp}

\subsection{Experimental Setup}
\label{sec::setup}

\textbf{Methods.} 
Recently, Chinese character recognition methods have achieved remarkable progress, showcasing impressive performance in general and zero-shot CCR. 
However, most evaluations are conducted using the CASIA-HWDB~\cite{hwdb} and CTW~\cite{ctw} datasets, which are severely constrained in the range of character categories and fall short in cultural heritage preservation and societal needs. 
To comprehensively evaluate character recognition methods on the mega-category scenario, we benchmark five types of methods on MegaHan97K, including general (ResNet~\cite{resnet}), radical sequence predicting-based (FewShotRAN~\cite{fewran}), radical embedding-based (HDE~\cite{hde}, RIE~\cite{rie}, SideNet~\cite{sidenet}, HierCode~\cite{hiercode}), image-IDS matching-based (CCR-CLIP~\cite{ccr_clip}), and glyph-based (OpenCCD~\cite{ccd}, PCSS~\cite{PCSS}) methods.
Stroke-based methods (SLD~\cite{sld}, ACPM~\cite{mmcharacter}) are not included due to the lack of stroke-level information.

These methods typically utilize interchangeable visual encoders, such as ResNet or ViT, to encode input images, while incorporating prior knowledge (e.g., strokes, radicals, and glyphs) of each Chinese character to capture its structural information. The primary distinction among these approaches lies in how they leverage the prior knowledge of characters.
\textbf{(1) Radical sequence predicting-based methods:} FewShotRAN~\cite{fewran} utilizes a visual encoder to extract character features, followed by a recurrent neural network to predict the IDS of Chinese characters. 
\textbf{(2) Radical embedding-based methods:} 
HDE~\cite{hde}, SideNet~\cite{sidenet}, and HierCode~\cite{hiercode} convert the IDS of Chinese characters into binary trees. HDE encodes node paths to generate unique character representations. SideNet enhances HDE with learnable parameters for adaptive feature adjustment during training. HierCode uses prototype encoding to traverse the binary tree for robust feature representation. RIE leverages radical entropy to construct unique character encodings.
After extracting visual features, these methods calculate the similarity between the visual features and the corresponding encoding table to produce the final predictions.
\textbf{(3) Image-IDS matching-based methods:} CCR-CLIP~\cite{ccr_clip} employs CLIP’s visual encoder and text encoder to encode character images and IDS respectively, aligning them within the feature space to further capture the relationships between visual and structural features of Chinese characters.
\textbf{(4) Glyph-based methods:} OpenCCD~\cite{ccd} achieves Chinese character recognition by comparing the features of the input image with the features of printed images. PCSS~\cite{PCSS} continuously updates the features of printed images through prototype learning, thereby obtaining more robust feature representations of Chinese characters.

\textbf{Implementation details.} 
For a fair comparison of HDE~\cite{hde}, RIE~\cite{rie}, and HierCode~\cite{hiercode}, we employ ResNet50~\cite{resnet} as the backbone and optimize these models using the Adam optimizer with initial $lr$=0.001, $\beta_{1}$=0.9, and $\beta_{2}$=0.99. 
The size of the input image is normalized to 96×96, and the batch size is set to 400. 
For SideNet~\cite{sidenet}, we only implement the DDCM version due to the extremely large memory demands of its complete version in the mega-category scenario. 
To handle the increased number of radicals from the extensive categories of Chinese characters, we expand the dimension of the radical prototype layer in FewShotRAN~\cite{fewran} and OpenCCD~\cite{ccd} from 512 to 1024.
Apart from these modifications, all remaining details are strictly implemented with their corresponding papers. 
Additionally, we reproduce RIE~\cite{rie}, while the other methods are implemented using either open-source code or code provided by the authors. 
We use character accuracy as the evaluation criteria. 
All experiments are conducted on a server equipped with an Intel Xeon Platinum 8360Y CPU @ 2.40GHz and an NVIDIA A6000 GPU with 48GB memory. The software environment consisted of Ubuntu 20.04 LTS, PyTorch 1.13.1, CUDA 11.7, and cuDNN 8.5.0.

\textbf{Dataset splitting.}
Following prior research~\cite{sld, rie, ccr_clip}, we conduct experiments in two settings: general and zero-shot CCR. 
An overview of the splitting of these two settings is included in Table~\ref{tab:dataset_div}. 
The general CCR setting indicates that the categories in the test set are included in the training set, as outlined in Table~\ref{tab::general_CCR_train} and Table~\ref{tab::general_CCR_test}. 
Conversely, in the zero-shot CCR setting, the categories in the test set are entirely absent from the training set. 
As detailed in Table~\ref{tab::zs_train} and Table~\ref{tab::zs_test}, we split the real data in MegaHan97K into two parts.
The 27,533 categories in the GB18030-2000 standard are used as training data, and the remaining 69,922 categories are used as testing data, ensuring that the training and test categories do not overlap.

\subsection{General Chinese Character Recognition}
\label{sec::general ccr}

Table~\ref{tab:generl_exp} and Table~\ref{tab::footp} demonstrate the results of general Chinese character recognition. 
We derive the following observations. 
\textbf{(1) The synthetic subset significantly enhances CCR performance.}
As shown in the last three columns in Table~\ref{tab:generl_exp}, it is evident that all models achieve remarkably superior performance when trained with the synthetic subset, exhibiting an impressive average improvement of 22.43\% compared to without the synthetic subset. 
This underscores the effectiveness of our synthetic subset and further highlights its utility as an efficacious solution to the challenge of collecting adequate training data in mega-category scenarios.
\textbf{(2) Mega-category scenarios present the challenge of exponentially increased storage requirements.}
As shown in Table~\ref{tab::footp}, the storage demands of all methods increase substantially under the mega-category setting, which is primarily attributed to the increment in the category number. 
The surge in categories not only increases storage demands in the classification layer but also significantly impacts the storage space needed for this prior information (e.g., structural, radical, and stroke information).
In comparison to 16,151 (the maximum number of categories in existing datasets), most models showcase larger than 60\% size increases in storage requirements under our 97,455 categories. 
It underscores that the mega-category nature of our dataset introduces novel challenges to CCR, particularly for computational resource requirements and the deployment of mega-category models on edge devices, such as phones and tablets.
This previously unforeseen observation offers novel insights that could spark new ideas for future research in the field of pattern recognition with the mega-category scenario.
\begin{table}[t]
\caption{Comparison of various methods in terms of average inference time (AIT), and accuracy. 
`w./o. syn.' and  `w. syn.' indicate without and with the synthetic subset, respectively. 
$\Delta$ denotes the accuracy improvement with the synthetic subset. 
The \textbf{bold} and \underline{underline} indicate the best and second best, respectively. 
}
\label{tab:generl_exp}
\centering
\renewcommand{\arraystretch}{1.0} 
\resizebox{1.0\linewidth}{!}{
\begin{tabular}{lcrrrr}
\specialrule{1.0pt}{1pt}{1pt}
\multirow{2}{*}[0ex]{Method} & \multirow{2}{*}[0ex]{Venue} & \multirow{2}{*}[0ex]{AIT $\downarrow$} 
& \multicolumn{2}{c}{ACC $\uparrow$ (\%)}  & \multirow{2}{*}[0ex]{$\Delta$ (\%)} \\ \cline{4-5}
              &                                                 & &w./o. Syn.         & w. Syn.            &            \\ 
\hline
ResNet50~\cite{resnet}        & CVPR’16    & 16ms            & 34.89              & 88.76              & +53.87  \\ 
FewShotRAN~\cite{fewran}      & PRL’19     & \textbf{12ms}          & 69.25              & 88.41              & +19.16   \\ 
HDE~\cite{hde}                & PR’20      & 15ms                   & 65.85              & 88.86              & +23.01   \\ 
OpenCCD~\cite{ccd}            & CVPR’22    & 16ms                  & \underline{79.49}  & 88.55              & +9.06    \\ 
RIE~\cite{rie}                & PR’23      & 15ms           & 62.92              & 88.36              & +25.44   \\ 
CCR-CLIP~\cite{ccr_clip}      & ICCV’23    & \underline{13ms}           & \textbf{82.04}     & \underline{89.56}  & +7.52    \\ 
SideNet~\cite{sidenet}   & PR’24      & 1809ms                   & 61.82              & 88.63              & +26.81   \\ 
HierCode~\cite{hiercode}      & PR’24   & 15ms              & 66.58              & \textbf{92.32}     & +25.74   \\  
PCSS~\cite{PCSS}      & ICASSP'24  & 14ms  & 78.27 & 89.53 & +11.26 \\
\specialrule{1.0pt}{1pt}{1pt}
\end{tabular}
}
\end{table}
\begin{table}[t]
\centering
\caption{Comparative analysis of model sizes, prior information sizes, footprints, and Additional Footprint Proportion (AFP) across various Chinese character recognition methods in the mega-category scenario.
`Prior size' refers to the size of additional prior information, such as structure and radical information. 
The footprint represents the sum of the model size and the size of the prior information.
AFP indicates the percentage increase in footprint when character categories expand from 16,151 to 97,455 relative to the total footprint.}
\label{tab::footp}
\renewcommand{\arraystretch}{1.0} 
\resizebox{1.0\linewidth}{!}{
\begin{tabular}{lcccc}
\specialrule{1.0pt}{1pt}{1pt}
Method          & Model Size & Prior Size & Footprint & AFP (\%) \\ \hline
ResNet50~\cite{resnet}      & 893.1MB    & -             & 893.1MB   &71.12 \\
FewShotRAN~\cite{fewran}     & 1735.6MB   & -                    & 1735.6MB  &79.19 \\
HDE~\cite{hde}             & 109.1MB    & 352.4MB                & 461.5MB   &63.62 \\
OpenCCD~\cite{ccd}       & 634.9MB    & 199.6MB                & 834.5MB  &19.95  \\
RIE~\cite{rie}             & 109.1MB    & 352.0MB                & \underline{461.1MB}   & 63.70  \\
CCR-CLIP~\cite{ccr_clip}      & 261.8MB    & 199.6MB                & 461.4MB  &36.09  \\
SideNet~\cite{sidenet}    & 83.3MB     & 10936.4MB              & 11019.7MB &82.79 \\
HierCode~\cite{hiercode}     & 109.0MB    & 174.3MB                & \textbf{283.3MB}  &51.32  \\
PCSS~\cite{PCSS}          & 30.6MB    & 1598.3MB            & 1628.9MB  & 81.86  \\
\specialrule{1.0pt}{1pt}{1pt}
\end{tabular}}
\end{table}
\textbf{(3) Morphologically similar and complex characters in MegaHan97K present significant challenges for CCR.} 
Ideographic Description Sequences (IDS) encapsulate the inherent structure of Chinese characters, comprised of radicals and structural symbols. 
We assume that Chinese characters with an IDS edit distance of no more than three are morphologically similar. 
Analysis of the edit distances between the IDS of erroneous samples and their corresponding ground truths reveals that 38.34\% of these distances are three or fewer.
This suggests that a substantial number of morphologically similar characters are misidentified, primarily due to the presence of the mega-category in the Chinese character lexicon.
Additionally, characters with complex shapes are also prone to confusion, with 74.18\% of the error samples featuring ten or more strokes.

\begin{table}[t]
\small
\caption{Comparison of various methods in zero-shot Chinese character recognition experiment. `Print.' denotes that the method uses additional printed images.
}
\label{zs_exp}
\centering
\renewcommand{\arraystretch}{1.0} 
\begin{tabular}{lccr}
\specialrule{1.0pt}{1pt}{1pt}
Method     & Venue & Print. & ACC $\uparrow$ (\%) \\ \hline 
FewShotRAN~\cite{fewran}      & PRL’19    &\ding{51}  & 46.76   \\ 
HDE~\cite{hde}                & PR’20     &\ding{55}  & 47.39            \\
OpenCCD~\cite{ccd}            & CVPR’22   &\ding{51}  & 76.06   \\ 
CCR-CLIP~\cite{ccr_clip}      & ICCV’23   &\ding{55}  & \textbf{79.04}   \\
RIE~\cite{rie}                & PR’23     &\ding{55}  & 45.41            \\
SideNet-DDCM~\cite{sidenet}   & PR’24     &\ding{55}  & \underline{47.68}\\ 
HierCode~\cite{hiercode}      & PR’24  &\ding{55}  & 46.82            \\ 
PCSS~\cite{PCSS}              & ICASSP'24   &\ding{51} &  77.34 \\
\specialrule{1.0pt}{1pt}{1pt}
\end{tabular}
\end{table}

\subsection{Zero-Shot Chinese Character Recognition}

We conduct zero-shot CCR experiments to further evaluate the zero-shot capability of existing methods.
The results listed in Table~\ref{zs_exp} reveal that: 
\textbf{(1) The mega-category of MegaHan97K poses a significant challenge for zero-shot CCR.} 
All methods deliver unsatisfactory performance under the mega-category setting, primarily due to the overly large category, which results in a significant number of radical zero-shot instances in the test set.
This introduces a scenario that is notably more complex than previous challenges. 
\textbf{(2) The CLIP-based method outperforms other approaches in the zero-shot scenario.} CCR-CLIP, an image-IDS matching-based method built upon CLIP, delivers the optimal zero-shot performance, surpassing its closest competitor, SideNet-DDCM, by 31.36\% in recognition accuracy. This superior performance may be attributed to its use of contrastive learning, which aligns image and text features in the same latent space, potentially enabling the model to better capture the structural information of Chinese characters.
\textbf{(3) Glyph-based method is effective in zero-shot recognition.} For methods that utilize additional printed character images for training or testing, OpenCCD outperforms FewShotRAN by achieving 29.3\% higher performance and demonstrates comparable performance to the optimal model CCR-CLIP. This effectiveness stems from the use of printed character templates, which can effectively introduce prior knowledge. However, these methods require a large number of printed images during training, leading to higher training costs.
\textbf{(4) Radical embedding-based methods have a smaller footprint but show limited effectiveness in the zero-shot scenario.} As shown in Tables 9 and 10, the radical embedding-based methods, such as HierCode, HDE, and RIE, have smaller footprints compared to other types of methods. However, this smaller footprint leads to a greater compression of prior information, which may result in information loss and subsequently degrade the model's performance.

\subsection{Long-tail distribution Problem Analysis}
\label{sec::long-tail}

We conduct experiments to analyze the long-tail distribution problem on the M$^5$HisDoc~\cite{m5} and MegaHan97K datasets.
Our assessment employs dual metrics: top-1 accuracy and macro accuracy~\cite{macro}, where macro accuracy is computed by aggregating individual category accuracies and calculating their arithmetic mean. 
The latter, particularly efficacious in quantifying model performance across the character distribution spectrum, shows high sensitivity to data-scarce categories~\cite{macro}. 
As shown in Table~\ref{tab::long-tail}, both HierCode~\cite{hiercode} and CCR-CLIP~\cite{ccr_clip} achieve high macro accuracy on the MegaHan97K dataset, indicating that the training set features a relatively balanced distribution of categories, enabling effective learning across all classes. 
Furthermore, the minimal discrepancy between top-1 and macro accuracy for both methods suggests that the test set of MegaHan97K is also well-balanced. 
Therefore, there is essentially no long-tail distribution problem in the MegaHan97K dataset. 
Conversely, both models demonstrate considerably lower macro accuracy relative to top-1 accuracy on the M$^5$HisDoc dataset, indicating severe long-tail distribution issues prevalent in both the training and test sets.

\begin{table}[t]
\centering
\caption{Comparative analysis of long-tail problem. Macro ACC is defined as calculating the accuracy for each category individually and then averaging the results. }
\label{tab::long-tail}
\renewcommand{\arraystretch}{1.0} 
\resizebox{0.98\linewidth}{!}{
\begin{tabular}{lcccccc}
\specialrule{1.0pt}{1pt}{1pt}
\multirow{2}{*}{Method}         & \multicolumn{2}{c}{MegaHan97K (Ours)} &  & \multicolumn{2}{c}{M$^5$HisDoc~\cite{m5}}  \\ \cline{2-3} \cline{5-6}
 & Top1 ACC $\uparrow$  & Macro ACC $\uparrow$ & & Top1 ACC $\uparrow$ & Macro ACC $\uparrow$  \\ \hline
CCR-CLIP~\cite{ccr_clip}      & 89.56\%    & 91.52\%   &  & 93.26\%    & 71.96\%   \\
HierCode~\cite{hiercode}      & 92.32\%    & 93.62\%   &  & 94.08\%    & 63.48\%   \\
\specialrule{1.0pt}{1pt}{1pt}
\end{tabular}}
\end{table}

\subsection{Investigating Synthetic Data Effects and Limitations}
\label{ablation}

In this subsection, we analyze the effects and limitations of synthetic data on model performance, summarized as follows:

\textbf{(1) Impact of Synthetic Data Volume:} We conduct an experimental study to investigate the effect of synthetic data volume, as depicted in Figure~\ref{fig::ablation}. 
With the addition of synthetic data, the accuracy first improves and tends to stabilize.
When the number of synthetic data samples per category is fewer than 35, accuracy exhibits an upward trajectory as additional data is incrementally introduced. 
However, once the number reaches approximately 35, accuracy plateaus at around 92.4\%, suggesting a point of diminishing returns. 
It is noteworthy that increasing the volume of synthetic data significantly prolongs the duration of training. 
Consequently, we select 35 samples per category for our training set to balance optimal accuracy with reasonable training time.

\textbf{(2) Limitations for Complex and Similar Characters:}
To verify whether increasing synthesized data for challenging samples can further improve model performance, we select Chinese characters with more than 10 strokes to represent complex characters and those with an IDS edit distance of less than 3 to represent similar characters. For these characters, we increase the number of synthesized samples to 70, while keeping the sample size at 35 for the remaining characters.
Despite this adjustment, experiments show only a marginal improvement in Top-1 accuracy, increasing from 92.32\% to 92.35\%.
Additionally, we use the latest state-of-the-art method IF-Font~\cite{iffont} to synthesize data and train the recognition model with 15 samples per category, achieving an accuracy of 90.96\%, comparable to FontDiffuser.

\begin{figure*}[t]
\centering
\includegraphics[width=0.85\textwidth]{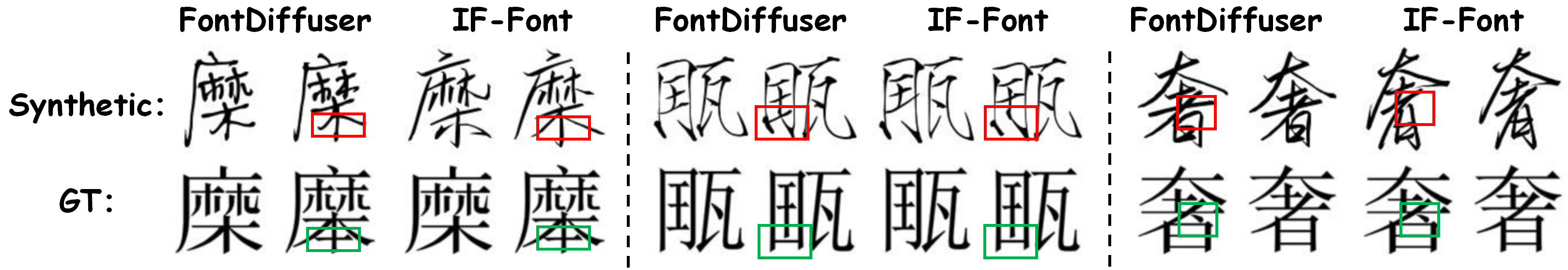}
\caption{Comparison of challenging samples synthesized by FontDiffuser and IF-Font with ground truth (GT). \textcolor{red}{Red} boxes indicate errors, while \textcolor{green}{green} boxes highlight correct components.}
\label{iffont}
\end{figure*}

\begin{figure}[t]
  \centering
  \vspace{-3mm}
  \includegraphics[width=0.95\linewidth]{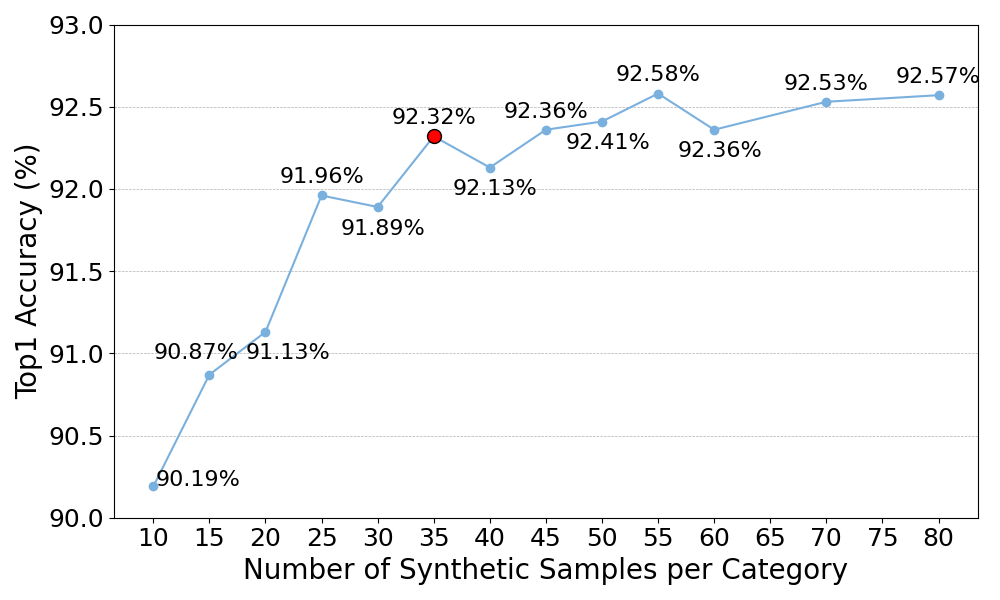}
  \caption{Effects of increasing synthetic sample sizes on accuracy.}
  \label{fig::ablation}
\end{figure}

To analyze the reason for such a marginal improvement, we perform a visualization of the synthesized samples for these complex and similar characters, as shown in Figure~\ref{iffont}. The results reveal that font synthesis models struggle with these challenging samples, exhibiting issues such as stroke misplacement and missing details. 
The incorrectly synthesized characters may confuse the recognition model, hampering the recognition accuracy. 
This indicates that merely increasing synthesized data for challenging samples or adopting more advanced models may still fail to address the challenges in mega-category Chinese character recognition.

\subsection{Cross-Validation with Other Datasets}
We also conduct cross-validation experiments on CASIA-AHCDB~\cite{ahcdb}, M$^5$HisDoc~\cite{m5}, CASIA-HWDB~\cite{hwdb}, and MegaHan97K using the CCR-CLIP~\cite{ccr_clip} model. Notably, the data from M$^5$HisDoc's test set are not included in MegaHan97K's training set.
Based on the results in Table~\ref{exp::cross}, we draw the following conclusions.
\textbf{(1) MegaHan97K manifests robust generalization capabilities.} The models trained on the MegaHan97K dataset exhibit robust generalization performance on other datasets.
\textbf{(2) The MegaHan97K dataset presents novel challenges to the field due to its inclusion of mega-categories.} 
The models trained on existing datasets exhibit underperformance on the MegaHan97K dataset. This is primarily attributed to the limited number of categories in existing datasets, which hinders the models' ability to generalize effectively to the more complex challenge posed by the mega-category scenario in the MegaHan97K dataset. 

\begin{table}[t]
\caption{Cross-validation of the CCR-CLIP~\cite{ccr_clip} model trained on other datasets and MegaHan97K. The entries in gray represent intra-dataset validations.}
\label{exp::cross}
\centering
\renewcommand{\arraystretch}{1.0} 
\resizebox{0.75\linewidth}{!}{
\begin{tabular}{cc}
\specialrule{1.0pt}{1pt}{1pt}
Task     & ACC $\uparrow$ (\%) \\ \hline 
CASIA-AHCDB → MegaHan97K &   12.05   \\ 
MegaHan97K → CASIA-AHCDB & \textbf{80.79}   \\ 
\textcolor{gray}{CASIA-AHCDB → CASIA-AHCDB} & \textcolor{gray}{94.23} \\ \hline 
CASIA-HWDB → MegaHan97K   & 20.65 \\ 
MegaHan97K → CASIA-HWDB &  \textbf{72.64}   \\  
\textcolor{gray}{CASIA-HWDB → CASIA-HWDB} & \textcolor{gray}{96.38} \\ \hline 
M$^5$HisDoc → MegaHan97K  & 29.12 \\ 
MegaHan97K → M$^5$HisDoc  & \textbf{75.34} \\ 
\textcolor{gray}{M$^5$HisDoc → M$^5$HisDoc} & \textcolor{gray}{93.26} \\ 
\specialrule{1.0pt}{1pt}{1pt}
\end{tabular}
}
\end{table}

\subsection{Analysis of the Impact of Combining MegaHan97K with Different Datasets}
\label{sec::comb}

In this subsection, we combine MegaHan97K with existing datasets for joint training to investigate its impact on the performance of these datasets. From the experimental results in Table~\ref{tab:combined_training_results} (Lines 1, 5, 6, 10, 11, and 12), it can be observed that integrating MegaHan97K with existing datasets significantly improves the model's Top-1 accuracy and Macro accuracy. The improvement in Macro accuracy indicates that the model's ability to recognize rare and variant characters has been enhanced, demonstrating greater robustness.

\begin{table}[t]
\centering
\caption{Performance comparison of CCR-CLIP model trained by combining MegaHan97K with different datasets. `Historical', `synthetic', and `handwritten' refer to the historical subset, synthetic subset, and handwritten subset, respectively.}
\label{tab:combined_training_results}
\resizebox{1.0\linewidth}{!}{
\begin{tabular}{clccc}
\hline
\textbf{\#Line} & \textbf{Training Set} & \textbf{Test Set} & \textbf{Acc $\uparrow$ (\%)} & \textbf{Macro Acc $\uparrow$ (\%)} \\ \hline
1 & HWDB                  & HWDB              & 96.38                  & 95.06                        \\
2 &HWDB+Historical                  & HWDB              & 92.60                  & 92.59                      \\
3 &HWDB+Synthetic                  & HWDB              & 96.67                  & 96.67                       \\
4 &HWDB+Handwritten                  & HWDB              & 96.74                  & 96.73                        \\
5 &HWDB+MegaHan97K          & HWDB              &  96.76             &  96.76                       \\ \hline
6 &AHCDB                 & AHCDB             & 94.23                  & 85.64                        \\
7 &AHCDB+Historical             & AHCDB             & 96.07                  & 92.08                        \\
8 &AHCDB+Synthetic             & AHCDB             & 97.91                  & 92.31                        \\
9 &AHCDB+Handwritten           & AHCDB             & 98.41                  & 94.21                        \\
10 &AHCDB+MegaHan97K      & AHCDB             & 98.49                  & 95.27                        \\ \hline
11 &M$^5$HisDoc              & M$^5$HisDoc          & 93.26                  & 71.96                        \\
12 &M$^5$HisDoc+MegaHan97K      & M$^5$HisDoc          & 95.37                  & 81.53                        \\ \hline
\end{tabular}
}
\end{table}

Furthermore, we select two representative datasets, HWDB (handwritten) and AHCDB (historical), to explore the impact of combining different MegaHan97K subsets (historical, synthetic, handwritten) during training on model performance.
The results in Table~\ref{tab:combined_training_results} show that for HWDB, the synthetic and handwritten subsets (lines 3 and 4) improve the model's performance. However, the historical subset (line 2) leads to a decrease in performance due to the significant stylistic differences between the historical subset and handwritten data.
In contrast, for AHCDB, all three subsets of MegaHan97K (lines 7–9 in Table~\ref{tab:combined_training_results}) positively contribute to model performance. This can be attributed to the data augmentation strategies applied to the synthetic and handwritten subsets in MegaHan97K (as illustrated in Figure~\ref{fig:data_process}), which effectively simulate the style of historical data.

\begin{figure*}[t]
\centering
\includegraphics[width=0.9\textwidth]{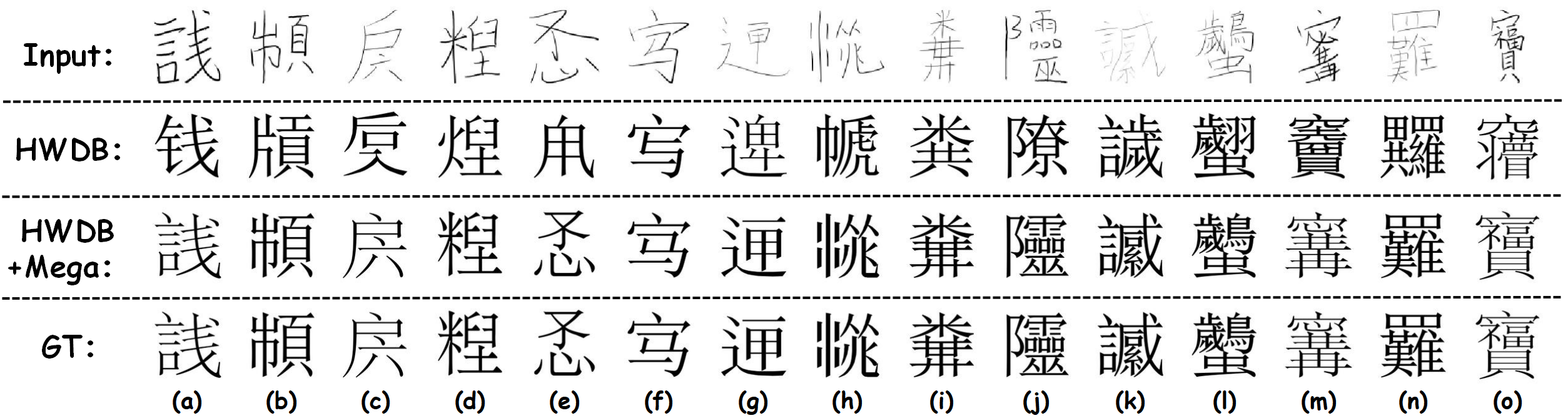}
\caption{Comparison of prediction results with and without the use of MegaHan97K. ``HWDB" represents predictions from the model trained only on the HWDB dataset, while ``HWDB+Mega" represents predictions from the model trained on both the HWDB and MegaHan97K datasets.}
\label{sidebyside}
\end{figure*}

To further analyze the advantages of combining MegaHan97K into model training, we use the HWDB dataset as an example and conduct a comparative visualization of the model outputs with and without MegaHan97K. As shown in Figure~\ref{sidebyside}, the characters that are incorrectly recognized without MegaHan97K but correctly identified when using it are mostly similar characters or rare characters. Whether it is relatively simple characters (e.g., Figure~\ref{sidebyside} (a)-(g)) or complex characters (e.g., Figure~\ref{sidebyside} (h)-(o)), the model performs poorly without the inclusion of MegaHan97K. The main reason for this phenomenon is the lack of such characters in the training data. 
Therefore, this analysis further highlights the significance of the proposed dataset in the field of Chinese character recognition, as it provides a mega-category of Chinese character data, including rare, visually similar, and complex characters.

\section{Discussion}
\label{sec::discussion}

Several existing datasets~\cite{m5, hwdb} contain annotations spanning three levels: character, text, and page levels. Although text-level and page-level data offer richer contextual semantic information that better reflects daily usage scenarios, character-level datasets maintain unique research value. Our focus on character-level recognition is motivated by the following factors: 

(1) Many of the characters within the 97K categories we collect are rare or variant forms, significantly complicating the collection of text line data that naturally includes these characters. Given the scarcity and specialized nature of these characters, character-level analysis provides a foundation for understanding these rare glyphs while enabling research into the evolution and variations of Chinese characters.

\begin{figure*}[t]
\centering
\includegraphics[width=0.9\textwidth]{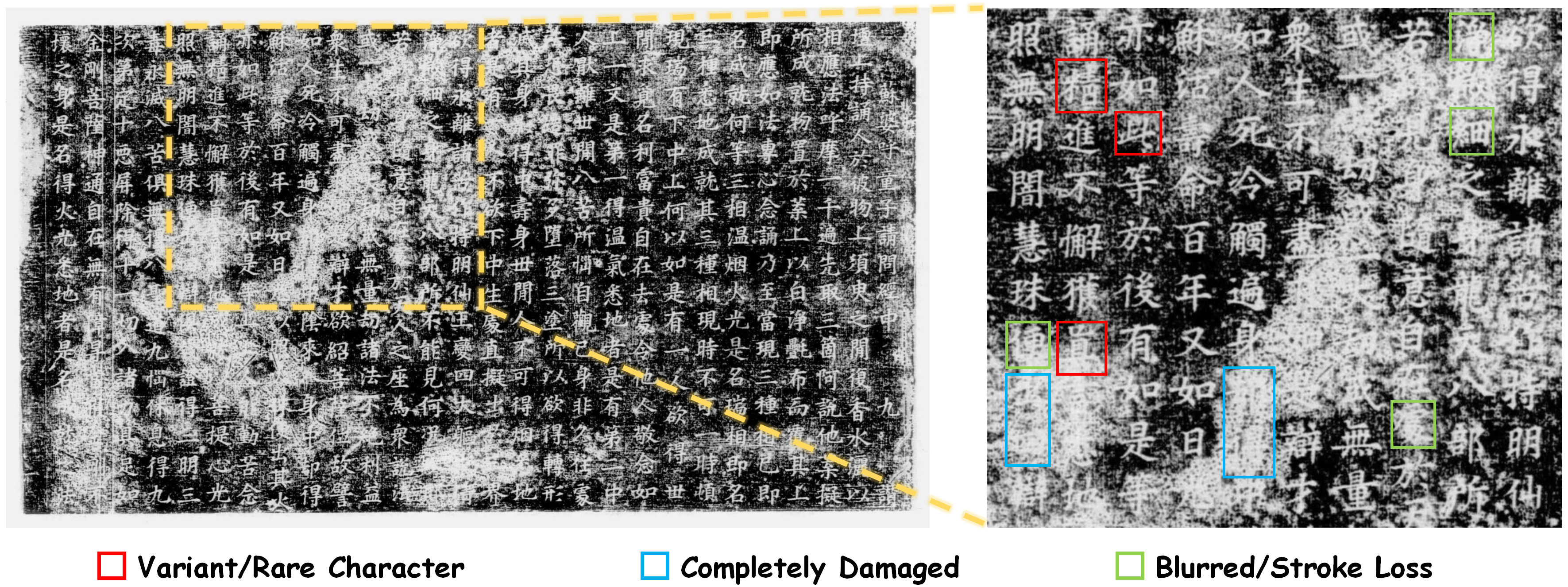}
\caption{Analysis of a damaged ancient document. The left side shows a sample of a damaged ancient document, while the right side provides a detailed qualitative analysis of specific regions. Highlighted cases include variant/rare characters (\textcolor{red}{red} boxes), completely damaged characters (\textcolor{blue}{blue} boxes), and blurred or stroke-loss characters (\textcolor{green}{green} boxes). Variant characters refer to characters that share the same meaning and pronunciation as their counterparts but differ in glyph, such as variations in radicals or strokes.}
\label{fssj}
\end{figure*}

(2) Our character-level dataset is particularly useful in scenarios such as the digitization and restoration of ancient documents~\cite{Dunhuang, diffhdr}. For example, as shown in Figure~\ref{fssj}, certain historical documents have suffered damage due to improper preservation, leading to the loss of contextual information and leaving only a few recognizable characters. In such cases, text-level recognition models may fail. Instead, a robust character-level recognition model capable of effectively identifying the remaining clear characters could assist historians in further digitization or restoration efforts. As demonstrated in Section~\ref{sec::comb} and Table~\ref{tab:combined_training_results} (Lines 6-12), the integration of our dataset with existing ancient datasets significantly improves the model's ability to recognize rare and variant characters, demonstrating greater robustness. Consequently, our dataset holds significant implications for specialized fields working with rare or ancient characters, particularly in the digitization and restoration of ancient documents.


\begin{figure*}[t]
\centering
\includegraphics[width=0.75\textwidth]{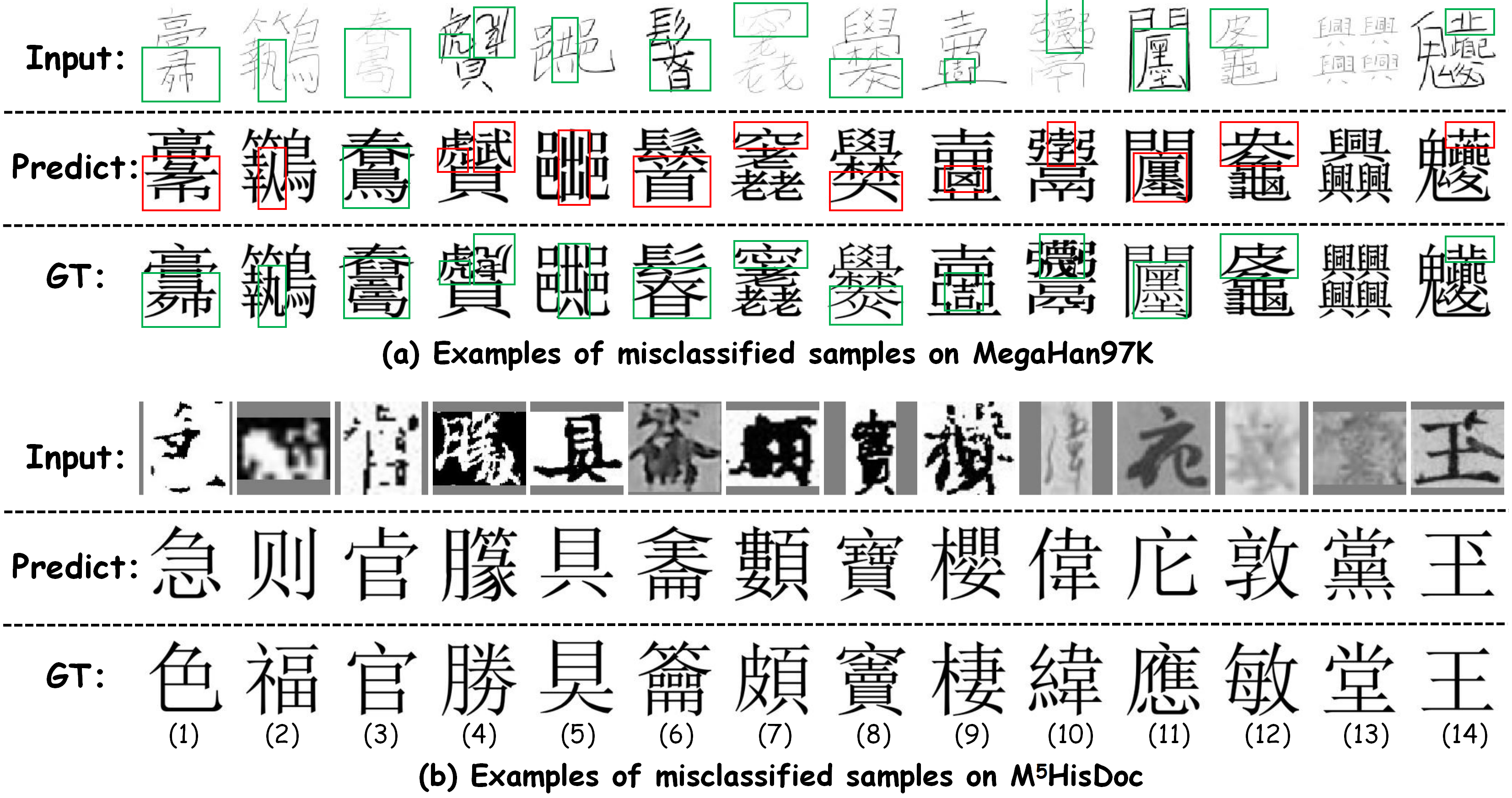}
\caption{Visualization analysis of the misclassified samples from the CCR-CLIP model. The misclassified components, such as radicals or strokes, are highlighted with \textcolor{red}{red} boxes, while the correctly recognized ones are marked with \textcolor{green}{green} boxes.}
\label{figure_badcase}
\end{figure*}

\section{Visualization}

In this section, we present visualizations of the misclassified samples from the best-performing CCR-CLIP model to further illustrate the challenges in mega-category Chinese character recognition.
As shown in Figure~\ref{figure_badcase} (a), this scenario contains a large number of complex Chinese characters, which are often highly similar to one another, differing only in subtle details such as strokes or individual radicals. 
Accurately identifying these characters requires models to have fine-grained perception of strokes and radicals, which presents a significant challenge for existing methods.
This observation further underscores the significance of our dataset, as it presents novel challenges for future research in large-vocabulary Chinese character recognition.

Additionally, we conduct experiments on the M$^5$HisDoc dataset to analyze potential challenging cases that might occur in real-world scenarios. As shown in Figure 10 (b), several issues in historical Chinese characters are identified, including missing strokes (Figure 10 (b) (1)-(3)), stroke adhesion (Figure 10 (b) (4)-(9)), illegible handwriting (Figure 10 (b) (10)-(11)), blurriness (Figure 10 (b) (12)-(13)), and extra strokes (Figure 10 (b) (14)). These issues result in misclassification and are worth further investigation in future research.

\section{Limitation}
\label{sec::limitation}


In this section, we analyze the limitations of the MegaHan97K and our strategies to address them as follows:

\textbf{(1) Coverage of Chinese Character Categories:} Despite our significant efforts to collect the widest possible range of Chinese character categories, there are some variant and rare categories that are not included in the current work due to their lack of standardized computer coding (the Chinese national standard~\cite{GB18030_2022} or Unicode\footnote{\href{https://home.unicode.org/about-unicode/}{https://home.unicode.org}}). If additional character categories are officially encoded in the future, we will consider expanding our dataset accordingly.

\textbf{(2) Domain Gap Between Writing on Tablets and Paper:} Although we thoroughly optimize the handwriting collection system to simulate writing on paper, we acknowledge a certain domain gap between writing on tablets and on paper. 
However, this gap has a limited impact on the model and can be mitigated with appropriate training strategies. For instance, as shown in lines 1 and 5 of Table~\ref{tab:combined_training_results}, combined training on the HWDB and MegaHan97K datasets slightly improves Top-1 accuracy and noticeable increases macro accuracy from 95.06\% to 96.76\%, demonstrating that combined training effectively reduces domain discrepancies and enhances model performance.

\textbf{(3) Limited Sample Size:} Due to a huge number of character categories, the collection and verification of the dataset are extremely labor-intensive and time-consuming. As a result, in the test set, we collect only five handwritten samples for each type of Chinese character. In the future, we plan to collect more samples to further expand the dataset.

\section{Conclusion}
In the paper, we introduce MegaHan97K, a mega-category, large-scale dataset that contains the largest 97,455 Chinese character categories. It is the first dataset that supports the latest Chinese GB18030-2022 standard and boasts the largest category to date, at least six times more expansive than existing datasets. 
We conduct an extensive benchmark evaluation of MegaHan97K and perform a detailed analysis of the results. 
Our findings suggest that MegaHan97K more accurately mirrors the challenges faced in cultural heritage preservation and societal needs, while also showing enhanced compatibility with contemporary Chinese character processing systems.
We believe this dataset will serve as an invaluable resource for researchers dedicated to advancing the field of mega-category Chinese character recognition.

\section*{Acknowledgement}
This research is supported in part by the National Natural Science Foundation of China (Grant No.:62476093, 62441604), and the National Key Research and Development Program of China  (2022YFC3301703).

\section*{}
\bibliography{egbib}

\end{document}